\documentclass{bmvc2k}

\title{WHENet: Real-time Fine-Grained Estimation for Wide Range Head Pose}

\addauthor{Yijun Zhou}{yijun.zhou1@huawei.com}{1}
\addauthor{James Gregson}{james.gregson@huawei.com}{1}

\addinstitution{
 IC Lab, Huawei Technologies Canada\\
}

\runninghead{Zhou, Gregson}{WHENet}

\def\etal{\emph{et al}\bmvaOneDot}

\newcommand{\first}[1]{\underline{\bf#1}}
\newcommand{\second}[1]{{\bf#1}} 
\usepackage{float}
\usepackage[title]{appendix}
\usepackage{hyperref}

\begin{document}

\maketitle

\begin{abstract}
We present an end-to-end head-pose estimation network designed to predict Euler angles through the full range head yaws from a single RGB image. Existing methods perform well for frontal views but few target head pose from all viewpoints. This has applications in autonomous driving and retail. Our network builds on multi-loss approaches with changes to loss functions and training strategies adapted to wide range estimation. Additionally, we extract ground truth labelling of anterior views from a current panoptic dataset for the first time. The resulting Wide Headpose Estimation Network (WHENet) is the first fine-grained modern method applicable to the full-range of head yaws (hence wide) yet also meets or beats state-of-the-art methods for frontal head pose estimation. Our network is compact and efficient for mobile devices and applications.  Code will be available at: \href{https://github.com/Ascend-Research/HeadPoseEstimation-WHENet}{https://github.com/Ascend-Research/HeadPoseEstimation-WHENet}

\end{abstract}

%-------------------------------------------------------------------------
\section{Introduction}
\label{sec:introduction}
Head pose estimation (HPE) is the task of estimating the orientation of heads from images or video (see Figure~\ref{fig:full_range_image}) and has seen considerable research. Applications of HPE are wide-ranging and include (but are not limited to) virtual \& augmented-reality~\cite{murphy2010head}, driver assistance, markerless motion capture~\cite{de2013robust} or as an integral component of gaze-estimation~\cite{murphy2008head} since gaze and head pose are tightly linked~\cite{langton2004influence}.

The importance of HPE is well described in~\cite{murphy2008head}. They describe wide-ranging social interactions such as {\em mutual gaze}, {\em driver-pedestrian} \& {\em driver-driver} interaction. It is also important in providing visual cues for the targets of conversation, to indicate appropriate times for speaker/listener role switches as well as to indicate agreement~\cite{murphy2008head}. For systems to interact with people naturally, it is important to be sensitive to head pose.

Most HPE methods target frontal to profile poses since applications are plentiful, the face is feature-rich and training datasets are widely available. However, covering full range is useful in many areas including driver assistance~\cite{murphy2007head}, motion-capture and to generate attention maps for advertising and retail~\cite{siriteerakul2012advance}.
\begin{figure*}[h!]
\begin{centering}
\includegraphics[width=0.8\textwidth]{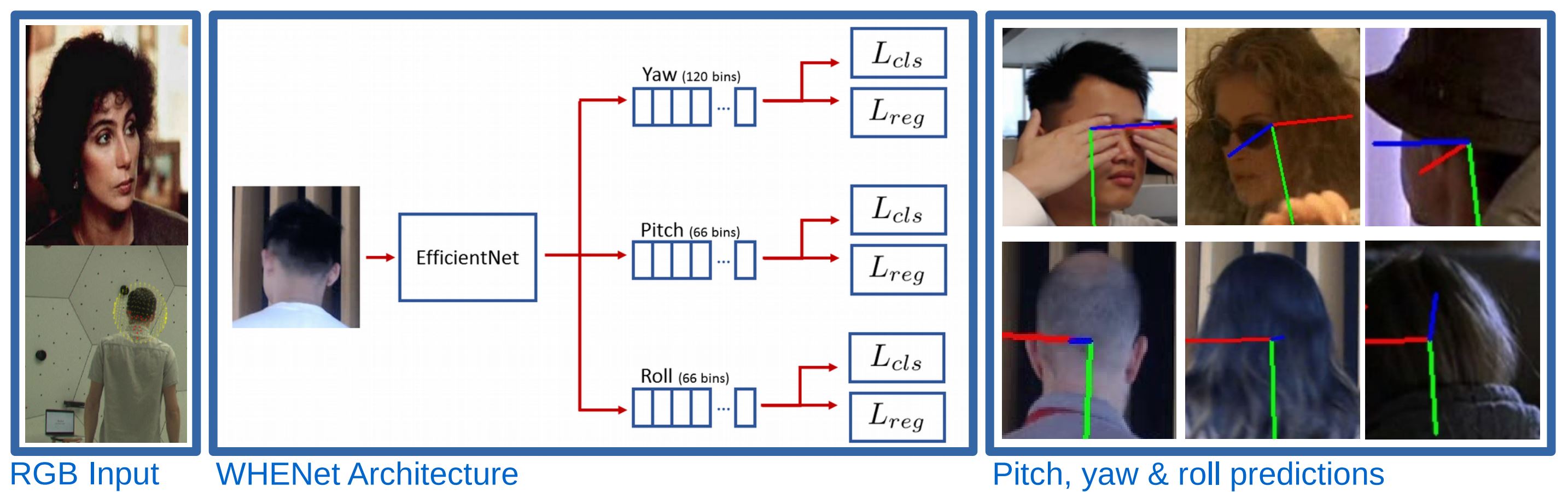}
\caption{The WHENet full-range head pose estimation network (center) refines~\cite{ruiz2018fine} by combining an EfficientNet\cite{tan2019efficientnet} backbone with classification and regression losses on each of pitch, yaw and roll. A new {\em wrapped-loss} stabilizes the network at large yaws. Training this network using anterior views repurposed from~\cite{joo2017panoptic} (lower-left) via a novel reprocessing procedure allows WHENet to predict head poses from anterior view as well as heads featuring heavy occlusions, fashion accessories and adverse lighting (right) with a mobile-friendly model. In spite of this, WHENet is as accurate or more accurate than existing state-of-the-art methods that are constrained to frontal or profile views. Some images are from~\cite{zhu2016face}\&\cite{vu2015context}}
\label{fig:teaser}
\end{centering}
\end{figure*}
        
Furthermore, though methods such as~\cite{ruiz2018fine,shao2019improving} perform well, they may be prohibitively large networks for mobile and embedded platforms. This is especially important in autonomous driving where many subsystems must operate concurrently. 

Based on these criteria, we developed the Wide Headpose Estimation Network (WHENet) which extends head-pose estimation to the full range of yaws (hence {\em wide}) using a mobile-friendly architecture.  In doing so, we make the following contributions:

\begin{itemize}
\item We introduce a {\em wrapped loss} that significantly improves yaw accuracy for anterior views in full-range HPE.
\item We detail an automated labeling process for the CMU Panoptic Dataset~\cite{joo2017panoptic} allowing it to be used for training and validation data in full-range HPE.
\item We extend our modified network (WHENet) to the full range of yaws where it achieves state-of-the-art performance for full-range HPE {\em} and within 1.8\% of state-of-the-art for narrow range HPE despite being trained for another task.
\item We demonstrate that simple modifications to HopeNet~\cite{ruiz2018fine} can achieve $29\%$ improvement over the original network and $5-13\%$ improvement over the current state-of-the-art FSANet~\cite{yang2019fsa} for narrow range HPE from RGB images in the AFLW2000~\cite{zhu2016face} and BIWI~\cite{fanelli_IJCV} datasets.
\end{itemize}
%In addition, by building on~\cite{ruiz2018fine} we retain the advantages of CNN-based HPE including robustness to occlusion \& fashion accessories, facial expressions, adverse lighting or poor image quality. These are common failure points for methods based on keypoints or morphable models.

\section{Background \& Related Work}
\label{sec:background} 
Head pose estimation has been actively researched over the past 25 years. Approaches up to 2008 are well surveyed in~\cite{murphy2008head} and much of our discussion on early methods follows their work. Due to the breadth of research activity and our target application of HPE from monocular RGB images, we exclude multi-view or active sensing methods from this review.

{\bf Classical methods} include template matching and cascaded detectors.  Template matching compares input images with a set of labeled templates and assign a pose based on nearby matches~\cite{sherrah1999understanding,ng2002composite}. They can have difficulty differentiating between identity and pose similarity~\cite{murphy2008head}. Cascaded detectors train distinct detectors that also localize heads for each (discretized) pose. Challenges include prediction resolution and resolving when multiple detectors fire~\cite{murphy2008head}. %Both approaches have been largely superseded though current state-of-the-art methods~\cite{ruiz2018fine,mukherjee2015deep,shao2019improving} use fine-grained classifiers to stabilize training. We adopt this approach with WHENet.

{\bf Geometric \& deformable models} are similar methodologies. Geometric models use features, e.g. facial keypoints, from the input images and analytically determine the matching head pose using a static template~\cite{burger2014self}. The majority of their complexity lies in detecting the features, which is itself a well studied problem, see e.g.~\cite{sun2013deep,herpers1996edge} and survey~\cite{wang2018facial}.  Deformable models are similar but allow the template to deform to match subject-specific head features. Pose or other information is obtained from the deformed model~\cite{yu2013pose,cai20103d,yang2002model,zhu2016face}.

{\bf Regression \& classification methods} serve as supersets or components of most other methods. Regression methods use or fit a mathematical model to directly predict pose based on labeled training data. The formulation of the regressor is wide-ranging and includes principle component analysis~\cite{zhu2004head,srinivasan2002head} and neural networks~\cite{ruiz2018fine,mukherjee2015deep} among others. In contrast, classification methods predict pose from a discretized set of poses. Prediction resolution tends to be lower, generally less than 10-12 discrete poses. Methods include decision trees \& random forests~\cite{benfold2008colour,fanelli2011real}, multi-task learning~\cite{yan2013no,yan2015multi} and neural networks~\cite{ruiz2018fine,mukherjee2015deep}. Our networks are most similar to the multi-loss framework in~\cite{ruiz2018fine} which uses classification and regression objectives. Other works employ soft stage-wise regression~\cite{yang2018ssr, yang2019fsa} by training with both classification and regression objectives at multiple scales.

{\bf Multi-tasks methods} combines the head pose estimation problem with other facial analysis problems. Studies\cite{chen2014joint, zhu2012face, ranjan2017hyperface,ranjan2017all} show that learning related tasks at the same time can achieve better performance than training tasks individually. For instance, in~\cite{ranjan2017hyperface,ranjan2017all} face detection and pose regression are trained jointly. 
 
{\bf Full-range methods} are much less common than narrow-range since most existing datasets for HPE focus on frontal to profile views. Recent methods include~\cite{raza2018appearance, heo2019estimation, rehder2014head} which classify poses into coarsely-grained bins/classes to determine yaw. Unlike our method, pitch and roll are not predicted in~\cite{raza2018appearance, heo2019estimation, rehder2014head}.

{\bf Datasets} for facial pose include BIWI~\cite{fanelli_IJCV}, AFLW2000\cite{zhu2016face} and 300W-LP~\cite{zhu2016face}. Both AFLW2000 and 300W-LP use a morphable model fit to faces under large pose variation and report Euler angles. 300W-LP generates additional synthetic views to enlarge the dataset. More recently, the UMD Faces\cite{bansal2017umdfaces} and Pandora\cite{borghi2017poseidon} datasets provide a range of data labels, including head pose. A disadvantage for our application is that they do not cover the full-range of head poses but only frontal-to-profile views.

Very important to our method is the CMU Panoptic Dataset~\cite{joo2017panoptic}. It captures subjects from an abundance of calibrated cameras covering a full-hemisphere and provides facial landmarks in 3D. Using this data, we are able to estimate head pose from near-frontal views and use this pose to label non-frontal viewpoints. By doing so, we cover the full range of camera-relative poses, allowing our method to be trained with anterior views. This is discussed further in Section~\ref{sec:dataset}

\section{Our Method}
\label{sec:method}
\begin{figure}
\centering
\subfigure[]{\label{fig:full_range_image}\includegraphics[width=.38\textwidth]{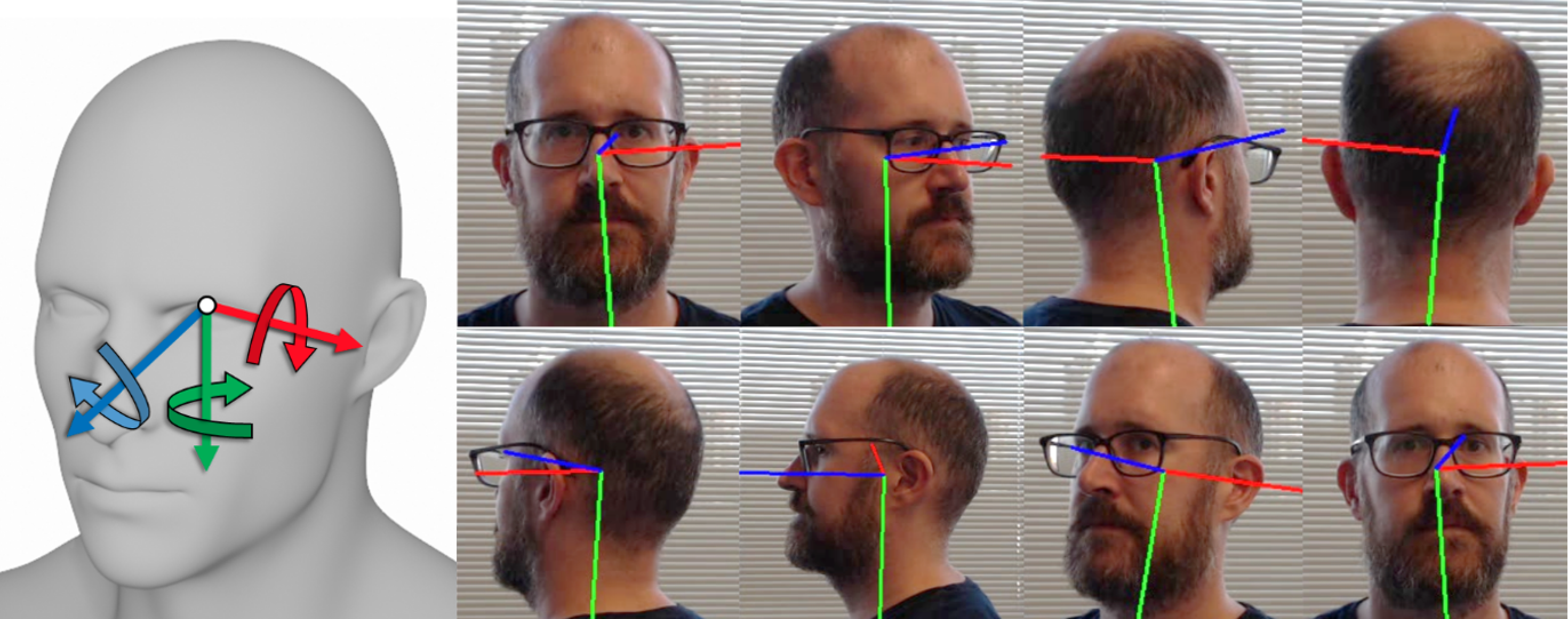}}
\hspace{0.5em}%
\subfigure[]{\label{fig:wrapped_loss}\includegraphics[width=.38\textwidth]{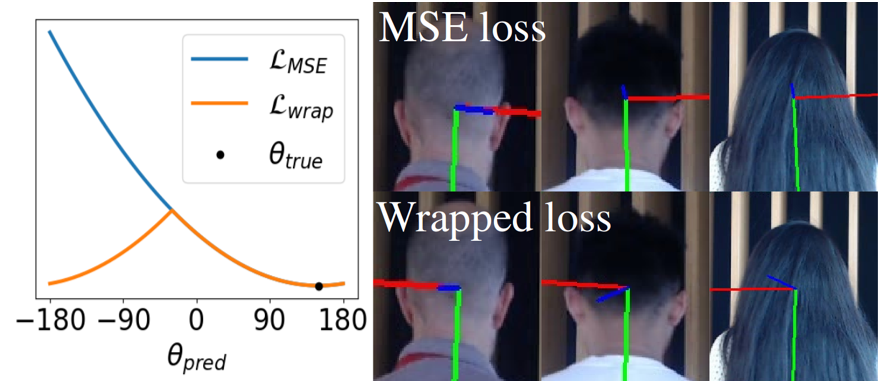}}
\hspace{1em}%
\subfigure[]{\label{fig:panoptic_frame}\includegraphics[width=.17\textwidth]{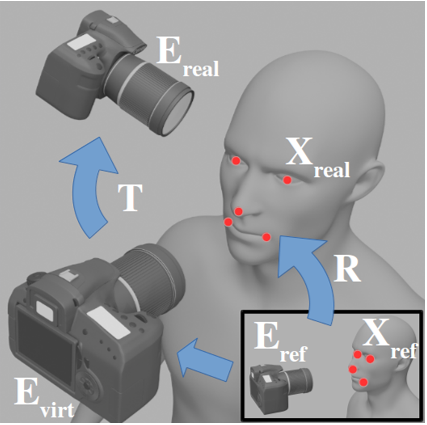}}

\caption{Head pose (a) is parameterized by pitch (red-axis), yaw (green-axis) and roll (blue-axis) angles in indicated directions. 
Our proposed wrapped loss function (b-left) avoids over-penalizing predictions $>180^\circ$ from true where pose is similar but MSE produces extreme loss values. This improves predictions when subjects face away from camera (b-right) where networks trained with MSE have errors approaching $180^\circ$ for yaw. Note the x (red) axis should align with the left ear of subjects. In (c), we compute virtual camera extrinsics oriented to provide a frontal view along with true extrinsics to extract Euler angles from the CMU Panoptic dataset~\cite{joo2017panoptic}. This allows us to automatically label tens of thousands of anterior view images for the full-range HPE task. We believe we are the first to   do so.}
\end{figure}
Our network design is derived from the multi-loss framework of Ruiz et al.~\cite{ruiz2018fine}. They combine a convolutional backbone with separate fully-connected networks that classify each of pitch, yaw and roll into $3^{\circ}$ bins using softmax with a cross-entropy loss.  A mean squared error (MSE) regression loss is also applied between the ground-truth labels and the expected value of the softmax output. The two losses are weighted to produce the final training objective for each angle. 

In~\cite{ruiz2018fine}, Ruiz \etal suggest that this combination of losses gives the resulting network robustness \& stability due to the softmax \& cross-entropy loss while still providing fine-grained supervision and output via the regression loss.

We adopt this overall framework from~\cite{ruiz2018fine} but make substitutions for both loss functions in order to adapt the method to full-range. To our knowledge we are the first to do so. Yaw prediction is divided into 120 $3^\circ$ bins covering the full range of yaws $(-180^\circ,180^\circ]$. Pitch and roll predictions are each made from 66 $3^\circ$ bins covering the range $[-99^\circ,99^\circ]$, although only the bins from $[-90,90]$ are ultimately used. The regression and classification losses for each of pitch, yaw and roll are combined via:
\begin{equation}
    \mathcal{L} = \alpha \mathcal{L}_{reg} + \beta \mathcal{L}_{cls}
\end{equation}
Here $\mathcal{L}_{cls}$ is the classification loss and $\mathcal{L}_{reg}$ is the regression loss while $\alpha$ and $\beta$ trade off the influence of one with respect to the other. 

%\subsection{Classification Loss}
We tested different classification losses and chose a sigmoid activation with binary cross-entropy for $\mathcal{L}_{cls}$. This differs from~\cite{ruiz2018fine} and shows marginally improved accuracy for the full-range task but is mentioned primarily for reproducibility.

%\subsection{Wrapped Regression Loss}
As in~\cite{ruiz2018fine} we predict output angles from the bin logits by applying softmax to obtain bin probabilities and take an expectation of the result for each of yaw, pitch and roll:
\begin{equation}
    \theta_{pred} = 3\sum_{i=1}^N p_i \left( i-\frac{1+N}{2} \right)
\end{equation}
Here $p_i$ is the probability of the i'th bin, $3$ is the bin width in degrees and $N$ is the bin-count of either $120$ (yaw) or $66$ (pitch and roll). The subtracted term shifts bin indices to bin centres.

In~\cite{ruiz2018fine}, a mean squared error (MSE) loss was used for regression which was sufficient for the limited range of yaws that were targeted. However, on the full-range task, this loss leads to erratic behaviour for subjects with absolute yaws exceeding $150^\circ$. This is illustrated in Figure~\ref{fig:wrapped_loss} and is due to $\pm180^\circ$ having wildly different angles for the same pose.

To prevent this, we define a {\em wrapped loss} (see Figure~\ref{fig:wrapped_loss}) that avoids this behavior. Rather than penalizing angle directly, it penalizes the minimal rotation angle that is needed to align each yaw prediction with its corresponding dataset annotation:

\begin{equation}
\mathcal{L}_{wrap}(\theta_{pred},\theta_{true}) = \frac{1}{N_{batch}} \sum_i^{N_{batch}} \min[|\theta^{(i)}_{pred}-\theta^{(i)}_{true}|^2, (360-|\theta^{(i)}_{pred}-\theta^{(i)}_{true}|)^2]
\end{equation}
%
%We then define the regression loss as the mean wrapped loss for the batch:
%
%\begin{equation}
%\mathcal{L}_{reg} = \frac{1}{N_{batch}} \sum_{i=1}^{N_{batch}} L_{wrap}( \theta^{(i)}_{pred},\theta^{(i)}_{true} )
%\end{equation}
%    
The loss function $\mathcal{L}_{wrap}$ is plotted in Figure~\ref{fig:wrapped_loss} for the target value $\theta_{true}=150^\circ$ and compared with an MSE loss. In the range $(-30,180]$ the two are identical but diverge for angles $<-30^\circ$. As they diverge MSE increases rapidly but the wrapped loss decreases as poses become more similar. The wrapped loss is smooth \& differentiable everywhere except at a cusp $180^\circ$ from the ground-truth yaw. This does make training the network more difficult, we suspect this is due to the cusp occurring at maximal errors in yaw.

The wrapped loss is key to the method's performance in anterior views, demonstrated empirically in Figure~\ref{fig:wrapped_loss}. A MSE trained network predicts entirely incorrect head poses while our predictions are consistent with the images. Figure~\ref{fig:histo_combine} plots errors as a function of angles and shows average errors at large yaw are decreased by more than 50\%.

In \cite{ranjan2017hyperface, ruiz2018fine} AlexNet and ResNet50 were used as backbones. These are quite large networks for a highly specialized task such as HPE. Since a focus of our method is to be mobile-friendly, we instead opted for a lighter backbone: EfficientNet-B0\cite{tan2019efficientnet}. EfficientNet-B0 is the baseline model of the EfficientNet family that incorporates Inverted Residual Blocks (from MobileNetV2\cite{sandler2018mobilenetv2}) to reduce the number of parameters while adding skip connections. 

This model size reduction is important on low-power embedded devices where head pose estimation may be only one component of a larger system. We have successfully ported a preliminary implementation to a low-power embedded platform where we see inference speeds approaching 60fps

\section{Datasets \& Training}
\label{sec:dataset}
Currently, major datasets for HPE are 300W-LP\cite{zhu2016face}, AFLW2000\cite{zhu2016face} and BIWI\cite{fanelli_IJCV}. Both AFLW2000 \& 300W-LP use 3D Dense Face Alignment (3DDFA) to fit a morphable 3D model to 2D input images and provide accurate head poses as ground-truth. 300W-LP additionally generates synthetic views, greatly expanding the number of images.

The BIWI dataset is composed of video sequences collected by a Kinect. Subjects moved their heads trying to span all possible angles observed from a frontal position. Pose annotations were created from the depth information. 

We follow the convention of~\cite{ruiz2018fine} by reserving AFLW2000~\cite{zhu2016face} and BIWI~\cite{fanelli_IJCV} for testing, while using 300W-LP~\cite{zhu2016face} for training.

Unfortunately, none of datasets mentioned above provide examples with (absolute) yaws larger than $100^\circ$. To overcome this, we generated a new dataset combining 300W-LP with data from the CMU Panoptic Dataset~\cite{joo2017panoptic}. The CMU Panoptic Dataset captured video of subjects performing tasks in a dome from approximately 30 HD cameras.

The panoptic dataset includes 3D facial landmarks and calibrated camera extrinsics and intrinsics but does not include head pose information. We use the landmarks and camera calibrations to locate and crop images of subjects' heads and to compute the corresponding camera-relative head pose Euler angles. We believe we are the first to use this dataset in this context.

Figure~\ref{fig:teaser} shows a frame from the dataset. The panoptic dataset has very little background variation and so cannot be used to train networks alone since networks do not learn features to differentiate subjects from general backgrounds. This is the motivation to combine it with 300W-LP, which is used for yaws in $(-99^\circ,99^\circ)$ while the panoptic dataset provides data mostly outside this range.
    
{\bf Processing the CMU Panoptic Dataset.} To compute camera-relative head pose Euler angles from the panoptic dataset we use the following procedure, depicted graphically in Figure~\ref{fig:panoptic_frame}. We first define a set of reference 3D facial landmarks ($x_{ref}$) matching the panoptic dataset facial keypoints annotations using a generic head model, except for the noisy jawline keypoints. A reference camera with intrinsics \& extrinsics ($K_{ref}$,$E_{ref}$) is then positioned to obtain a perfectly frontal view (yaw=pitch=roll=0) of $x_{ref}$. For each subject in each frame, we estimate the rigid transformation $R$ between $x_{ref}$ and the true keypoints $x_{real}$ provided by the panoptic dataset using~\cite{horn1988closed}. We then construct new extrinsics for a virtual camera $E_{virt} = E_{ref} R^{-1}$. This provides a (nominally) frontal view of the subject as they are positioned within the dome. Using the known real camera extrinsics $E_{real}$ from the panoptic dataset and $E_{virt}$, we recover the rigid transformation $T = E_{real} E_{virt}^{-1}$ between the virtual camera extrinsics and each real camera. Finally we extract Euler angles in pitch-yaw-roll (x-y-z) order from $T$, mirroring yaw and roll in order to match the rotation convention from the datasets. There is a translation component to the rigid transform but it is not needed to determine the head orientation with respect to the camera optical axes.

Using the synthetic reference and virtual camera in this process ensures that we have a consistent and automatic method for labeling. It also considerably reduces noise compared to manually selecting a frontal camera and annotating corresponding Euler angles.

To crop images, we define a spherical set of points around the subject's head as a {\em helmet} and project these into each view to determine a cropping region. Inspired by Ming et al.\cite{shao2019improving}, we leave a margin from the head bounding box so the network can learn to distinguish foreground and background. We followed the method of \cite{shao2019improving}, using K = 0.5 for data from 300W-LP and adjust the helmet radius to 21cm for CMU data.

{\bf Training} of WHENet is devided into two stages. We first train a narrow range model, WHENet-V with prediction range between $[-99^\circ, 99^\circ]$ for yaw, pitch and roll. We train this narrow range network on the 300W-LP~\cite{zhu2016face} dataset using an ADAM optimizer with learning rate 1e-5. Full-range WHENet with 120 yaw bins is then trained starting from the WHENet-V weights using our full-rage data set combining 300W-LP and CMU Panoptic Dataset data. We use the same optimizer and learning rate for this step. This two-stage approach helps the full-range network converge better and learn more useful features since the CMU data has little background variation. During training, images are randomly downsampled by up to 15X to improve robustness.

Training data from 300W-LP and the panoptic dataset is used in nearly equal amounts, with the former providing narrow range samples and the latter primarily handling wide range. A small amount of panoptic data is also used to level a dip in the histogram of narrow range images near yaw=0.

\section{Results \& Discussion} 
\label{sec:results}
For wide range results in this section, we define the Absolute Wrapped Error (AWE) as ${AWE} = \min\left(|\theta_{pred}-\theta_{true}|,360-|\theta_{pred}-\theta_{true}| \right)$ which properly handles wrapping of yaws.  We also define Mean AWE (MAWE) as the arithmetic mean of AWE. For results below, we are using $\alpha$ = 1, $\beta=1$ for WHENet and $\alpha=0.5$, $\beta=2$ for WHENet-V based on our ablations. Hyperparameter details can be found in our supplementary.
%Quantitative wide range results (in MAWE) for our method are summarized in Table~\ref{tab:summary_results} using our hybrid dataset and training strategy described in Sections~\ref{sec:dataset} \& \ref{sec:training} respectively.

Table~\ref{tab:summary_results} summarizes key results from WHENet and WHENet-V. We compare with eight narrow \& two full-range methods and report, where applicable/available, the output range, number of parameters, mean average errors on BIWI~\cite{fanelli_IJCV}, AFLW-2000~\cite{zhu2016face}, an average of both (to indicate generalization) and reported full-range MAE if applicable. WHENet \& WHENet-V are trained on 300W-LP~\cite{zhu2016face} or our combined dataset and are not trained on the AFLW2000 or BIWI datasets, as is standard practice. We begin with a discussion of the full-range network WHENet, as this is our primary application.

\begin{table}[h]
\centering
\caption{Summary of results: `Aggr. MAE': average of BIWI and AFLW2000 overall results. `Full MWAE': full-range results. \first{first/only}, \second{second}, -: not reported}
\setlength{\tabcolsep}{2pt}
\begin{tabular}{l |c c c c c c}
\hline
Method                                               & Full        & Params              & BIWI               & AFLW2k           & Avg.                     & Full          \\
                                                         & Range     & ($\times10^6$) & MAE$^\circ$    & MAE$^\circ$    & MAE$^\circ$           & MAWE$^\circ$   \\
\hline
KEPLER~\cite{kumar2017kepler}            & N            &  -                  & 13.852             & -                      & -                           & -           \\
Dlib (68 points)~\cite{kazemi2014one}                    & N            &     -                   & 12.249             & 15.777             & 14.013                   & -           \\
FAN (12 points)~\cite{bulat2017far}                        & N            &  6-24               & 7.882               & 9.116               &  8.499                   & -           \\
3DDFA~\cite{zhu2016face}                    & N            &  -                     & 19.07               & 7.393               & 13.231                  & -           \\ 
Shao \etal(K=0.5)~\cite{shao2019improving}     & N            &  24.6               & 5.999               & 5.478               &  5.875                   & -           \\
Hopenet~\cite{ruiz2018fine}                  & N             &  23.9               & 4.895               & 6.155               &  5.525                   & -           \\
SSR-Net-MD~\cite{yang2018ssr}          & N             & \first{0.2}           & 4.650               & 6.010               &  5.330                   & -            \\
FSA-Caps-Fusion~\cite{yang2019fsa}                 & N             & \second{1.2}           & 4.000               & \second{5.070} & \second{4.535}      & -            \\    
Rehder et al.~\cite{rehder2014head}    & Yaw         &  -                & -                      & -                      &  -                          & 19.0   \\  
Raza et al.~\cite{raza2018appearance}  & Yaw         &  -                    & -                      & -                      &  -                          & \second{11.25} \\  
\hline   
WHENet-V                                         & N            & 4.4    & \first{3.475}       & \first{4.834}     & \first{4.155}         & - \\
WHENet                                             & Y             & 4.4    & \second{3.814}  & 5.424               & 4.619                   & \first{7.655} \\
\end{tabular}
\label{tab:summary_results}
\end{table}
{\bf Full-range WHENet results and comparisons} are shown in Table~\ref{tab:summary_results}. Of the three full-range methods compared, WHENet is the only method to also predict pitch and roll. Comparisons are difficult to perform objectively since both Raza et al.~\cite{raza2018appearance} and Rehder et al.~\cite{rehder2014head} predict yaws only (no pitch and roll) and do so on non-public datasets. For~\cite{raza2018appearance}, MAE was not reported, so we report the lowest possible errors given a uniform distribution of yaws with their bin-widths (i.e. the result if their method performed perfectly). WHENet still shows a 31\% improvements, although these results should be treated as qualitative given lack of consistent testing data.  This latter point is something we attempt to address in this work by making our dataset processing code for the CMU Panoptic Dataset~\cite{joo2017panoptic} public.

Additionally, excluding WHENet-V, full-range WHENet has the lowest average errors on BIWI, second-lowest errors on AFLW2000 and has second lowest average overall errors, missing first place by 0.084$^\circ$ (1.8\%) to FSANet~\cite{yang2019fsa}. This is significant since full-range WHENet was not trained specifically for the narrow range task, yet is still state-of-the-art or very competitve with FSANet~\cite{yang2019fsa}, a significantly more complex network. In handling full-range inputs, WHENet is significantly more capable than FSANet, yet sacrifices remarkably little accuracy.

Figure~\ref{fig:full_range_image} shows pose predictions generated using WHENet to track a subject rotating through a full revolution of yaw. WHENet produces coherent predictions throughout, even when the face is completely absent. A key to achieving this is our wrapped loss function, which significantly improves anterior views. Without the wrapped loss predictions are erratic and often nearly 180$^\circ$ from the true pose as shown in Figure~\ref{fig:wrapped_loss}. This improvement is enabled by our annotation process for the CMU panoptic dataset~\cite{joo2017panoptic} which allows us to automatically generate labelings for tens of thousands of anterior views to tackle this challenge.

{\bf Narrow range results and comparisons} are listed in Tables~\ref{tab:narrow}. Dlib\cite{kazemi2014one}, KEPLER\cite{kumar2017kepler} and FAN\cite{bulat2017far} are all landmark based methods. 3DDFA\cite{zhu2016face} uses CNN to fit a 3D dense model to a RGB image. Hopenet\cite{ruiz2018fine} and FSANet\cite{yang2019fsa} are two landmark-free methods both of which explore the possbility of treating a continuous problem (head pose) into different classes/stages. As showed in Tables~\ref{tab:narrow}, WHENet-V achieves state-of-the-art accuracy on both BIWI~\cite{fanelli_IJCV} and AFLW2000~\cite{zhu2016face}, and outperforms the previous state-of-the-art FSANet~\cite{yang2019fsa} by $0.52^\circ$ (13.1\%) and $0.24^\circ$ (4.7\%) respectively, leading to an aggregate improvement of 0.38$^\circ$ (8.4\%) on the two datasets overall. It also achieve first place in every metric. Interestingly, WHENet-V shows a 29\% and 21\% improvement over HopeNet~\cite{ruiz2018fine} on these datasets, despite of having a very similar network architecture. %WHENet-V beating the current state-of-the-art demonstrates the strength of the general framework introduced by Ruiz et al.~\cite{ruiz2018fine}.
%
%Table~\ref{tab:AFLW_full_range} summarizes results for the AFLW2000\cite{zhu2016face} dataset and Table~\ref{tab:biwi} summarizes the same for the BIWI~\cite{fanelli_IJCV} dataset. Again, neither dataset was used for training. WHENet-V serves as an approximate lower bound for prediction errors in the narrow range case since full-range WHENet is targeting a wider prediction range.
\begin{table}
\centering
\caption{Comparison with state-of-the-art on BIWI\cite{fanelli_IJCV}(left) and AFLW2000\cite{zhu2016face}(right) dataset. Best (bold underlined) and second best (bold) results are highlighted.}
\begin{tabular}{l|llll|llll}
\hline
                & \multicolumn{4}{c}{BIWI}   & \multicolumn{4}{|c}{AFLW2000}\\
\hline
Method          & Yaw  & Pitch & Roll & MAE  & Yaw   & Pitch  & Roll & MAE  \\
\hline
KEPLER~\cite{kumar2017kepler}     & 8.80  & 17.3 & 16.2 & 13.9 & -     & -     & -     & -     \\
Dlib (68 points)~\cite{kazemi2014one}              & 16.8 & 13.8 & 6.19  & 12.2 & 23.1 & 13.6 & 10.5 & 15.8\\
FAN (12 points)~\cite{bulat2017far}            & -     & -     & -     & -     & 8.53  & 7.48  & 7.63  & 7.88  \\
3DDFA~\cite{zhu2016face}            & 36.20 & 12.30 & 8.78  & 19.10 & 5.40  & 8.53  & 8.25  & 7.39  \\
Hopenet(a = 1)~\cite{ruiz2018fine}  & 4.81  & 6.61  & 3.27  & 4.90  & 6.92  & 6.64  & 5.67  & 6.41  \\
Hopenet(a = 2)~\cite{ruiz2018fine}  & 5.12  & 6.98  & 3.39  & 5.12  & 6.47  & 6.56  & 5.44  & 6.16  \\
Shao \etal(K=0.5)~\cite{shao2019improving}     & 4.59  & 7.25  & 6.15  & 6.00  & 5.07  & 6.37  & 4.99  & 5.48  \\
SSR-Net-MD ~\cite{yang2018ssr}      & 4.49  & 6.31  & 3.61  & 4.65  & 5.14  & 7.09  & 5.89  & 6.01  \\
FSA-Caps-Fusion~\cite{yang2019fsa} & 4.27  & 4.96  & \second{2.76}  & 4.00  & \second{4.50}  & \second{6.08}  & \second{4.64}  & \second{5.07}  \\
\hline
WHENet-V        & \first{3.60}  & \first{4.10}  & \first{2.73}  & \first{3.48}  & \first{4.44}  & \first{5.75}  & \first{4.31}  & \first{4.83}  \\
WHENet          & \second{3.99}  & \second{4.39}  & 3.06  & \second{3.81}  & 5.11  & 6.24  & 4.92  & 5.42 
\end{tabular}

\label{tab:narrow}
\end{table}

%\todo{Add error histograms for narrow range WHENet-V}
%\begin{figure}[h]
%\begin{centering}
%\includegraphics[width=0.45\textwidth]{images/AFLW_hist_2.png}
%\caption{Histogram of the errors on AFLW2000 for WHENet-V and WHENet.}
%\label{fig:histo_aflw2000}
%\end{centering}
%\end{figure}

%\subsection{Loss Functions}
%\label{sec:loss}
\begin{figure}[h]
\begin{centering}
\caption{Histogram of the errors on our hybrid dataset for WHENet and WHENet (MSE), errors at high yaw are signficantly reduced with our proposed wrapped loss.}
\includegraphics[width=0.9\textwidth]{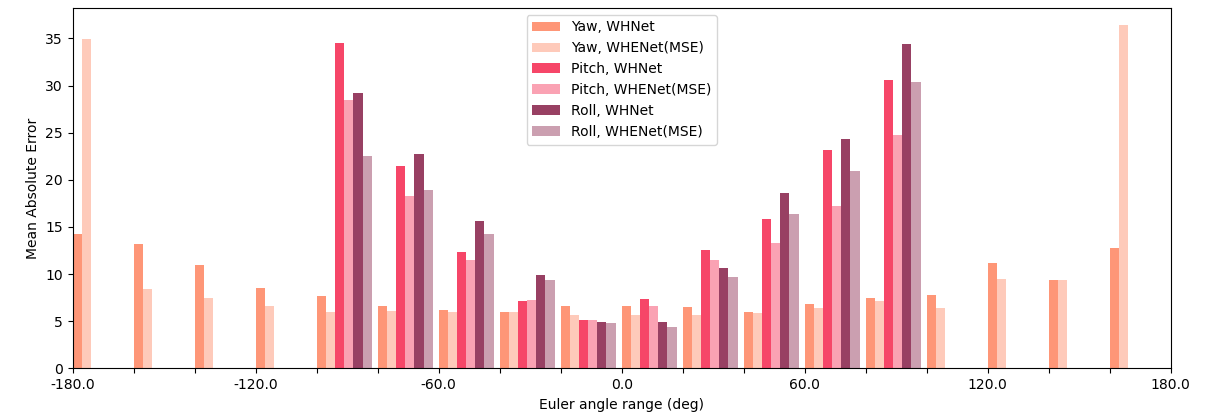}

\label{fig:histo_combine}
\end{centering}
\end{figure}
{\bf Ablation studies} for the change of backbone and loss functions are shown in Figure~\ref{fig:histo_combine} and Table~\ref{tab:loss_ablation_mse_wrapped}. Figure~\ref{fig:histo_combine} plots the mean errors at different angle intervals between wrapped regression loss (WHENet) and MSE loss (WHENet MSE).The mean error in yaw for extreme poses (close to -180/180 degrees) reaches 35 degrees when using MSE loss. By introducing the wrapped loss, errors for these angles are reduced by more than 50\% and are more consistent with those from medium or low yaws. High pitch and roll errors are typical in HPE methods as yaws approach $\pm90^\circ$ as seen in~\cite{shao2019improving}, we believe due to gimbal lock in the data labeling. Additionally, a quantitative result comparison can be found in Table \ref{tab:loss_ablation_mse_wrapped}. WHENet CE uses cross-entropy loss for classification and wrapped regression loss for regression. WHENet MSE uses the binary cross-entropy loss for classification and MSE loss for regression. WHENet uses binary cross-entropy loss for classification and wrapped regression loss for regression. We can see in the wide-range dataset (combine), WHENet has superior performance in yaw which confirms our observations in Figure~\ref{fig:histo_combine}. Although WHENet sacrifices some performance in the narrow range testing compared with other two loss settings, our main focus is in the wide-range predictions where we see significantly improved errors at large yaws. We believe performance could be improved by adapting the template keypoints used in annotating the CMU dataset to each subject rather than using a fixed template but leave this to future work.
%Table~\ref{tab:loss_ablation} compares a baseline method WHENet(MSE $\beta=1$) trained with an MSE regression loss.  Switching to the wrapped loss-function for WHENet($\beta=1$) improves the yaw accuracy by $1.2^\circ$, but worsens pitch and roll by $0.44^\circ$ and $0.63^\circ$ respectively. Tuning the tradeoff parameter to $\beta = 2$ gives our final network with slightly degraded pitch and roll accuracy ($0.05^\circ$ and $0.23^\circ$ respectively) but improves yaw accuracy by $1.9^\circ$. Overall the mean error is improved by $0.54^\circ$ however the bulk of this improvement comes for anterior views with absolute yaws exceeding $150^\circ$ which account for approximately $\frac{1}{6}$'th of the dataset. The qualitative improvement for such views is shown in Figure~\ref{fig:wrapped_loss}.

\begin{table}[h]
\centering
\caption{Result comparison of different loss settings. Combine indicates our combined dataset of CMU panoptic and 300W-LP}
\scalebox{0.8}{%
\begin{tabular}{l|llll|llll|llll}
\hline
           & \multicolumn{4}{c}{Combine}  & \multicolumn{4}{|c}{BIWI}   & \multicolumn{4}{|c}{AFLW2000} \\
\hline
           & Yaw   & Pitch & Roll & MWAE   & Yaw  & Pitch & Roll & MAE  & Yaw   & Pitch  & Roll & MAE  \\
\hline
WHENet CE  & 8.75  & 7.65  & 6.74  & 7.71  & \first{3.51} & \first{4.13}  & \first{3.04 }& \first{3.56} & 5.50  & 6.36   & 4.94 & 5.60 \\
WHENet MSE & 9.69  & \first{7.15}  & \first{6.19}  & 7.68  & 3.79 & 4.82  & 3.21 & 3.94 & \first{5.03}  & 6.41   & 5.16 & 5.53 \\
\hline
WHENet     & \first{8.51} & 7.67 & 6.78 & \first{7.66} & 3.99 & 4.39  & 3.06 & 3.81 & 5.11  & \first{6.24}   & \first{4.92} & \first{5.42}
\end{tabular}}

\label{tab:loss_ablation_mse_wrapped}    
\end{table}

The supplement provides additional ablation studies on the metaparameters $\alpha$ and $\beta$ as well the effects of resolution and relative comparison to video and depth-based methods that are outside our target application.

\section{Conclusions and future work}
In this paper we have presented WHENet, a new method for HPE that can estimate head poses in the full 360 degree range of yaws. This is achieved by careful choice of our wrapped loss function as well as by developing an automated labeling method for the CMU Panoptic Dataset~\cite{joo2017panoptic}. We believe we are the first to adapt this dataset to the specific task of head-pose estimation.  WHENet meets or exceeds the performance of state-of-the-art methods tuned for the specific task of frontal-to-profile HPE in spite of being trained for the full range of yaws. We are not aware of competing methods with similar capabilities and accuracy. 

%We further showed that WHENet is robust to a wide range of real-world issues including hairstyles, facial occlusions, fashion accessories, low-resolutions, poor cropping of heads and adverse imaging conditions. Finally we concluded with examples autonomous driving and driver monitoring systems that incorporate head pose to help gage pedestrian awareness of the vehicle and driver attention to improve safety.

In the future, we would like to extend upon this work by reducing network size even further. Reducing input image resolution has the potential to lower memory usage as well as allow shallower networks with fewer features. This could yield even further improvements in speed and size of network.

Another interesting avenue is modifying the representation of head pose. Euler angles are minimal and interpretable but have the drawback of gimbal lock. The pitch-yaw-roll rotation ordering of existing datasets such as AFLW2000 and BIWI emphasize this effect at yaws near $\pm90^\circ$ where pitch and roll effectively counteract each other. We believe this leads to the relatively high pitch and roll errors near profile poses seen in Figure~\ref{fig:histo_combine} which can also be seen in methods such as Shao et al.~\cite{shao2019improving}. Relabeling data to yaw-pitch-roll order might reduce this. Similarly other rotation representations such as axis-angle, exponential map or quaternions may help, though adapting the architecture to alternative rotation representations is likely non-trivial due to the classification networks.
\section{Acknowledgment}
We thank Shao Hua Chen and Tanmana Sadhu for their insightful discussions. We also thank Zhan Xu, Peng Deng, Rui Ma, Ruochen Wen, Qiang Wang and other colleagues in Huawei Technologies for their support in the project as well as the anonymous reviewers whose comments helped to improve the paper.
\bibliography{egbib}

\begin{appendices}
\section{Hyperparameter Studies}

Tables~\ref{tab:aflw_meta_ablation}, \ref{tab:biwi_meta_ablation}, \ref{tab:aflw_meta_ablation_w}, \ref{tab:biwi_meta_ablation_w} and \ref{tab:hybrid_meta_ablation_w} show ablation studies of mean average error for the $\beta$ and $\alpha$ metaparameters of WHENet-V and WHENet, tested on the AFLW2000, BIWI datasets and our combined dataset. From this, we selected the best overall performance as $\beta=2$ and $\alpha=0.5$ for WHENet-V, $\beta=1$ and $\alpha=1$ for WHENe, although performance is not overly sensitive to these choices. 

\begin{table}[h!]
\caption{WHENet-V MAE vs. $\alpha$ and $\beta$ on AFLW2000}
\centering
\begin{tabular}{l c c c}
\hline
                   & $\alpha=0.5$ & $\alpha=1$ & $\alpha=2$ \\
\hline
$\beta=0.5$ &  4.984            &    4.966     &   5.113      \\
$\beta=1$    &  4.946            &    5.146     &   4.904      \\
$\beta=2$    &  4.834            &    4.953     &   5.189      
\end{tabular} 
\label{tab:aflw_meta_ablation}
\end{table}

\begin{table}[h!]
\caption{WHENet-V MAE vs. $\alpha$ and $\beta$ on BIWI}
\centering
\begin{tabular}{l c c c}
\hline
                   & $\alpha=0.5$ & $\alpha=1$ & $\alpha=2$ \\
\hline
$\beta=0.5$ & 3.531             &    3.501    &  3.554      \\
$\beta=1$   &  3.551             &    3.676    &  3.626      \\
$\beta=2$   &  3.475             &    3.466    &  3.513      
\end{tabular} 
\label{tab:biwi_meta_ablation}
\end{table}

\begin{table}[h!]
\caption{WHENet MAE vs. $\alpha$ and $\beta$ on AFLW2000}
\centering
\begin{tabular}{l c c c}
\hline
                   & $\alpha=0.5$ & $\alpha=1$ & $\alpha=2$ \\
\hline
$\beta=0.5$ &  5.822            &    5.624     &   5.620      \\
$\beta=1$    &  5.484            &    5.424     &   5.529      \\
$\beta=2$    &  5.658            &    5.414    &   5.509      
\end{tabular} 
\label{tab:aflw_meta_ablation_w}
\end{table}

\begin{table}[h!]
\caption{WHENet MAE vs. $\alpha$ and $\beta$ on BIWI}
\centering
\begin{tabular}{l c c c}
\hline
                   & $\alpha=0.5$ & $\alpha=1$ & $\alpha=2$ \\
\hline
$\beta=0.5$ & 3.823             &    3.855    &  3.880      \\
$\beta=1$   &  3.843             &    3.814    &  3.786      \\
$\beta=2$   &  3.710             &    4.064    &  3.935      
\end{tabular} 
\label{tab:biwi_meta_ablation_w}
\end{table}

\begin{table}[h!]
\caption{WHENet MAE vs. $\alpha$ and $\beta$ on our combined dataset}
\centering
\begin{tabular}{l c c c}
\hline
                   & $\alpha=0.5$ & $\alpha=1$ & $\alpha=2$ \\
\hline
$\beta=0.5$ & 8.009             &    8.287    &  7.394      \\
$\beta=1$   &  7.331             &    7.655    &  7.879      \\
$\beta=2$   &  7.878             &    7.457    &  7.694      
\end{tabular} 
\label{tab:hybrid_meta_ablation_w}
\end{table}

\begin{table}[]
\caption{Comparison results on BIWI dataset with different modality methods. WHENet and WHENet-V are trained on 300W-LP and our combined dataset. The rest of the methods are trained on BIWI where they split the BIWI dataset into testing and training.}
\centering
\begin{tabular}{lllll}
\hline
                & Yaw  & Pitch & Roll & MAE  \\
\hline
\multicolumn{5}{l}{\bf{RGB-based}}                \\
\hline
DeepHeadPose~\cite{mukherjee2015deep}    & 5.67 & 5.18  & -    & -    \\
SSR-Net-MD~\cite{yang2018ssr}      & 4.24 & 4.35  & 4.19 & 4.26 \\
VGG16~\cite{gu2017dynamic}           & 3.91 & 4.03  & 3.03 & 3.66 \\
FSA-Caps-Fusion~\cite{yang2019fsa} & 2.89 & 4.29  & 3.60 & 3.60 \\
WHENet-V        & 3.60 & 4.10  & 2.73 & 3.47 \\
WHENet          & 3.99 & 4.39  & 3.06 & 3.81 \\
\hline
\multicolumn{5}{l}{\bf{RGB+Depth}}                \\
\hline
DeepHeadPose~\cite{mukherjee2015deep}    & 5.32 & 4.76  & -    & -    \\
Martin~\cite{martin2014real}          & 3.6  & 2.5   & 2.6  & 2.9  \\
POSEidon+~\cite{borghi2018face}       & 1.7  & 1.6   & 1.7  & 1.6  \\
\hline
\multicolumn{5}{l}{\bf{RGB+Time}}                 \\
\hline
VGG16+RNN~\cite{gu2017dynamic}       & 3.14 & 3.48  & 2.6  & 3.07
\end{tabular}
\end{table}

\section{Robustness}
A key objective of WHENet is to be robust to adverse imaging conditions as well as occlusions and accessories such as eyewear and hats. Much of the robustness of WHENet can be derived from using a similar network architecture as Hopenet~\cite{ruiz2018fine} which also performs well due to the CNN architecture.

\begin{figure}[h]
\begin{centering}
\includegraphics[width=.6\textwidth]{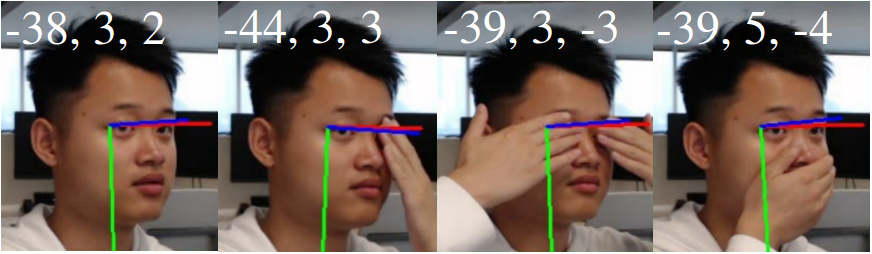}
\caption{Head pose estimation with occlusion. Subjects asked to remain still while covering different regions of their face. Predicted deviations are within $7^\circ$ of the unoccluded view (left). Some amount of deviation is expected due to slight subject motions.}
\label{fig:occlusion}
\end{centering}
\end{figure}

 Figure~\ref{fig:occlusion} shows a selection of occluded face images where the subject tried to maintain consistent head pose while blocking areas of their face. The angular predictions are quite stable with angles varying by only $7^\circ$ in spite of siginificant occlusions of features (some underlying variation of pose is expected due to subject motion).  This suggests the method is learning high-level features rather than specific localized details.

\begin{figure}[h]
\begin{centering}
\includegraphics[width=.6\textwidth]{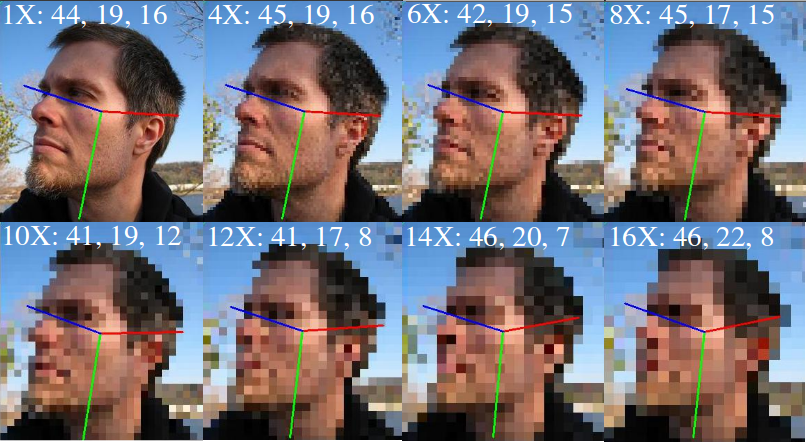}
\caption{Downsampling factor vs. yaw, pitch \& roll. Ground-truth values were 47.6, 22.0, 18.8. Images were downsampled by indicated amount and then resized to their original size using nearest-neighbor interpolation before being supplied to WHENet. Head pose predictions remain relatively stable event when images are aggressively downsampled by up to 16X. Original image from~\cite{zhu2016face}}
\label{fig:resolution_example}
\end{centering}
\end{figure}

We also evaluted the effect of resolution.  Figure~\ref{fig:resolution_example} illustrates qualitatively that prediction accuracy is not seriously degraded by aggressive downsampling of up to 16X.

\begin{figure}[h]
\begin{centering}
\includegraphics[width=0.55\textwidth]{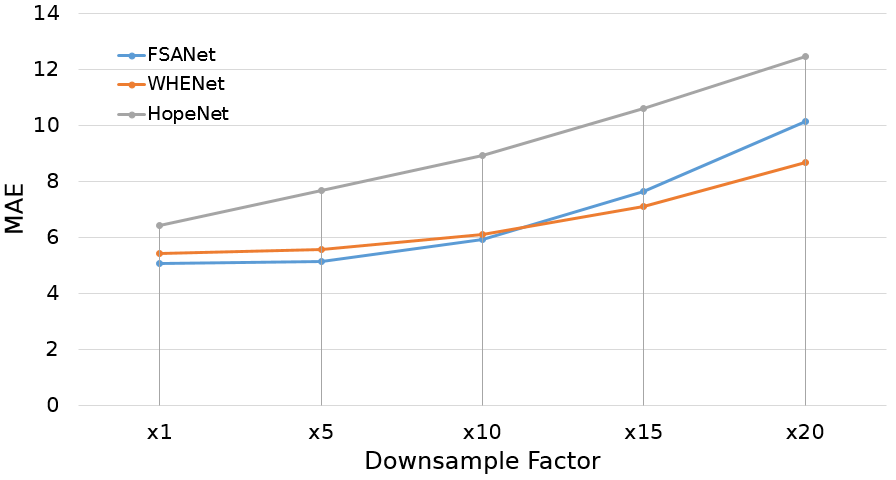}
\caption{Effect of downsampling factor on MAE. WHENet (orange) shows consistent improvement over the already impressive Hopenet~\cite{ruiz2018fine} and FSANet~~\cite{yang2019fsa} performance (grey and black). For Hopenet we plot the minimum (best) value at each downsampling factor among all training strategies reported in~\cite{ruiz2018fine}}
\label{fig:resolution_study}
\end{centering}
\end{figure}

We carried out this test in aggregate on the AFLW2000 dataset. The results are shown in Figure~\ref{fig:resolution_study} and compared to Hopenet~\cite{ruiz2018fine} and FSANet~~\cite{yang2019fsa}. We list the smallest reported errors for Hopenet among the four training strategies in~\cite{ruiz2018fine} and thank the authors for providing this data.

In summation, full-range WHENet targets a task that is outside the scope of the existing state-of-the-art using a faster and significantly smaller network. In spite of this, it meets or beats state-of-the-art performance for the restricted case of HPE for frontal-to-profile views when evaluated on two datasets that were not used during training.
\begin{figure}[h]
\centering
\includegraphics[width=0.7\textwidth]{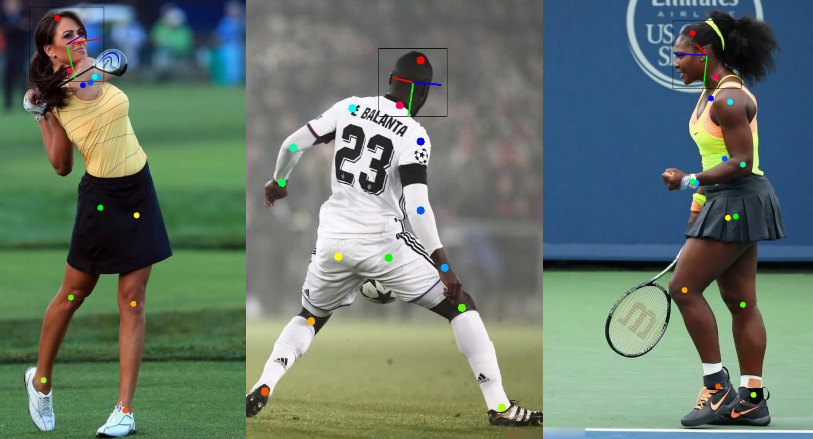}
\caption{WHENet applied to head crops generated from keypoint predictions from~\cite{osokin2018real}, keypoints shown as dots, illustrating how HPE can be integrated with full-body pose estimation methods. Images from~\cite{wu2017ai}}
\label{fig:results_openpose}
\end{figure}
\section{Applications}
\label{sec:applications}
Here we show qualitative examples of WHENet applied to several applications that demonstrate how HPE can integrate with real-world systems and how our training strategy allows the method to generalize to low-resolution and low-quality data that was not present during training.

Figure~\ref{fig:results_openpose} shows using a pose detector based on Lightweight OpenPose~\cite{osokin2018real} code to detect pose keypoints while using WHENet to predict head pose. Frequently pose-estimations do not estimate sufficient keypoints for accurate HPE but by incorporating a full-range HPE method such as WHENet, such limitations may be overcome. This could be used, for example, in sports broadcasting or by coaching staff to estimate participants fields of views and situational awareness when analyzing plays.

\begin{figure}[h]
\centering
\includegraphics[width=.7\textwidth]{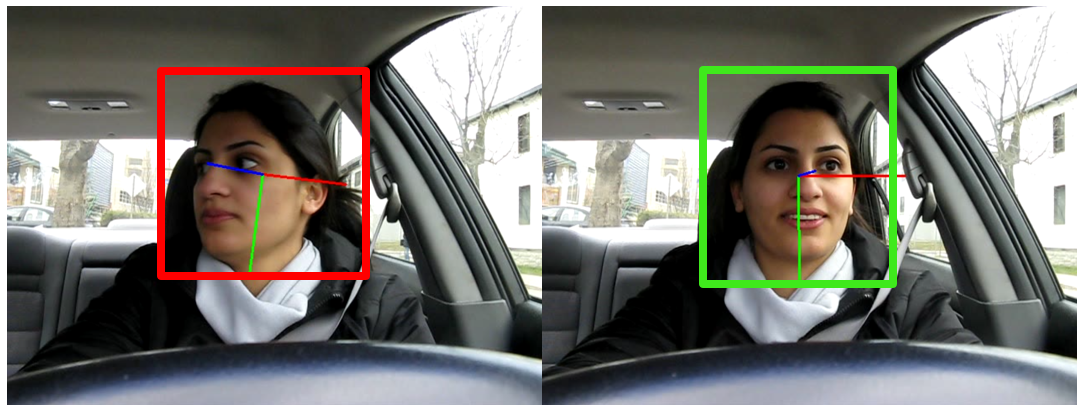}
\caption{Applications to autonomous driving and driver assistance. Left: Green boxes indicate yaws $<\pm45^\circ$ and potential awareness of vehicle, red boxes indicate probable inattention. This example highlights the need for efficient and low-resolution approaches to HPE with 6 total low-resolution detections. Here low-quality pose-estimates yield poor cropping regions but WHENet successfully generalizes despite having no comparable training data. Images from~\cite{geiger2013vision}. Right: WHENet is used to monitor driver attention, marking the driver as inattentive (red) when yaw exceeds $30^\circ$ and attentive (green) otherwise. Images from~\cite{abtahi2014yawdd}}
\label{fig:kitti_results}
\end{figure}

Figure~\ref{fig:kitti_results} depicts a hypothetical driver-attention module where drivers are considered attentive with camera-relative yaws $<30^\circ$ and inattentive otherwise. The extension to full-range could extend this to predicting blind spots during other activities such as reversing without requiring additional hardware.

\end{appendices}
\end{document}